\documentclass[10pt,twocolumn,letterpaper]{article}

\usepackage{wacv}              % To produce the CAMERA-READY version
\usepackage{amsmath,amsfonts}
\usepackage{algorithmic}
\usepackage{algorithm}
\usepackage{array}
\usepackage{textcomp}
\usepackage{stfloats}
\usepackage{url}
\usepackage{verbatim}
\usepackage{graphicx}

\usepackage{booktabs}
\usepackage{multicol,multirow}
\usepackage{bbold}
\usepackage{enumitem}
\usepackage{amssymb}
\usepackage[table]{xcolor}
\newcommand{\green}[1]{\textcolor{ForestGreen}{#1}}

\definecolor{wacvblue}{rgb}{0.21,0.49,0.74}
\usepackage[pagebackref,breaklinks,colorlinks,allcolors=wacvblue]{hyperref}

\def\wacvPaperID{1916} % *** Enter the WACV Paper ID here
\def\confName{WACV}
\def\confYear{2026}

\title{PrevMatch: Revisiting and Maximizing Temporal Knowledge\\in Semi-Supervised Semantic Segmentation}

\author{
Wooseok Shin \quad
Hyun Joon Park \quad
Jin Sob Kim \quad
Juan Yun \quad
Se Hong Park \quad
Sung Won Han\thanks{Corresponding author.}\\
Department of Industrial and Management Engineering, Korea University\\
{\tt\small \{wsshin95, winddori2002, jinsob, yunjuan, sehong1996, swhan\}@korea.ac.kr}
}

\begin{document}
\maketitle

\begin{abstract}
In semi-supervised semantic segmentation, the Mean Teacher- and co-training-based approaches are employed to mitigate confirmation bias and coupling problems. However, despite their high performance, these approaches frequently involve complex training pipelines and a substantial computational burden, limiting the scalability and compatibility of these methods. In this paper, we propose a PrevMatch framework that effectively mitigates the aforementioned limitations by maximizing the utilization of the temporal knowledge obtained during the training process. The PrevMatch framework relies on two core strategies: (1) we reconsider the use of temporal knowledge and thus directly utilize previous models obtained during training to generate additional pseudo-label guidance, referred to as previous guidance. (2) we design a highly randomized ensemble strategy to maximize the effectiveness of the previous guidance.
PrevMatch, a simple yet effective plug-in method, can be seamlessly integrated into existing semi-supervised learning frameworks with minimal computational overhead. Experimental results on three benchmark semantic segmentation datasets show that incorporating PrevMatch into existing methods significantly improves their performance. Furthermore, our analysis indicates that PrevMatch facilitates stable optimization during training, resulting in improved generalization performance.
\end{abstract}

\section{Introduction}
\label{sec:intro}
%% The Goal of Semi Seg (SSS) task
Semantic segmentation is a critical task in various computer vision-related applications, such as autonomous driving \cite{wang2020deep,xie2021segformer}, medical image analysis \cite{fan2020pranet,shin2022comma}, and robotics \cite{milioto2019bonnet}, where the goal is to assign a semantic class label to each pixel in an image.
Despite the recent success of supervised learning-based methods in semantic segmentation, obtaining precise pixel-level annotations for supervised learning is extremely time-consuming and expensive.
This limits the applicability of supervised learning-based methods across various domains and fields.
Thus, extensive research has been conducted in semi-supervised semantic segmentation to overcome this limitation. The research has focused on developing methods enabling models to learn effectively from a limited number of labeled images along with a large number of unlabeled images.

%% Describe SSS core concept
In semi-supervised learning, self-training \cite{lee2013pseudo,arazo2020pseudo,chen2022debiased, zhou2020time, yang2022st++} and consistency regularization \cite{sajjadi2016regularization,laine2016temporal,miyato2018virtual,xie2020unsupervised,chen2021semi,yang2023revisiting} have become the predominant approaches for utilizing unlabeled data.
In particular, self-training involves generating pseudo-labels using the predictions of the current model at each iteration for unlabeled samples and leveraging them to train the model in conjunction with labeled data.
Consistency regularization encourages a network to predict consistently for various perturbed forms of identical input.
Recent studies have focused on designing frameworks that combine self-training and consistency regularization to exploit the strengths of each method.
%% Fundamental Problem
However, self-training-based methods still suffer from the confirmation bias problem \cite{arazo2020pseudo}, where the model becomes overfitted in the wrong direction, even when combined with consistency regularization.

\begin{figure*}[ht]
\centering
\includegraphics[width=0.99\linewidth]{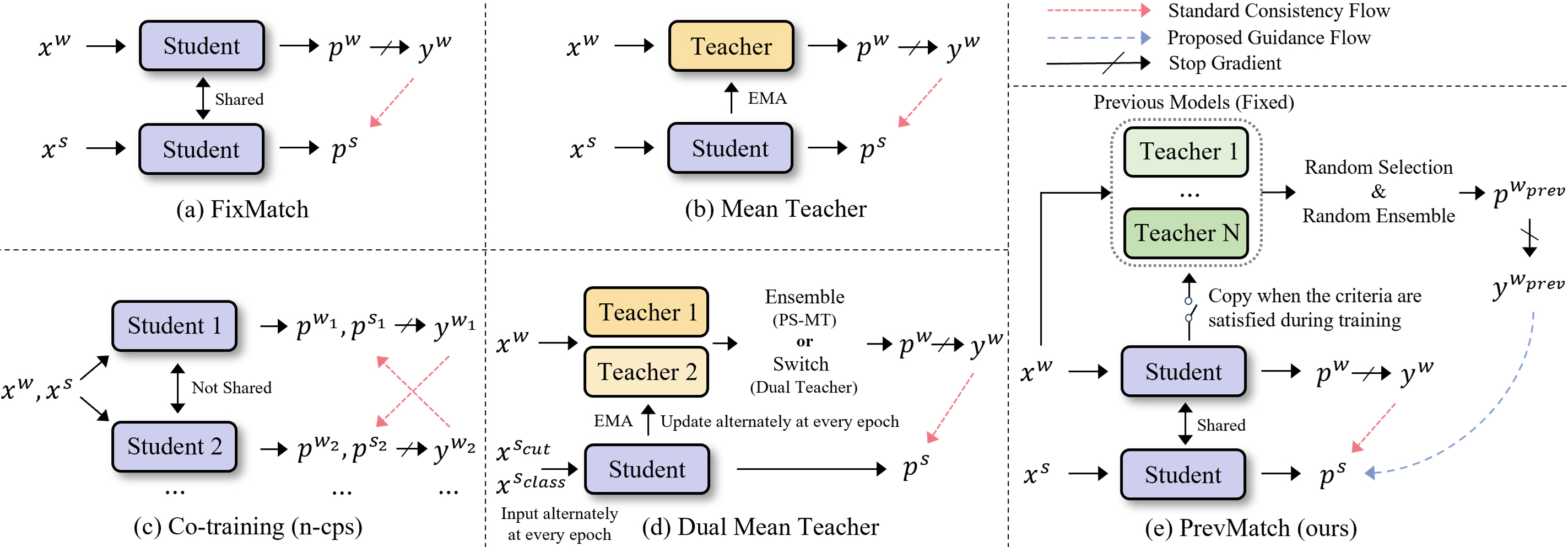}
\caption{Illustration of the frameworks for (a) FixMatch \cite{sohn2020fixmatch}, (b) Mean Teacher-based structure \cite{tarvainen2017mean}, (c) Co-training \cite{chen2021semi,li2023diverse} (cps: cross pseudo supervision), (d) Dual Mean Teacher \cite{liu2022perturbed, na2023switching}, and (e) the proposed method. In (d), the inputs ($x^{s_{cut}}, x^{s_{class}}$) indicate the CutMix \cite{yun2019cutmix} and ClassMix \cite{olsson2021classmix} augmentations used in Dual Teacher \cite{na2023switching}.}
\label{fig1:framework_comparison}
\end{figure*}

%% Existing Methods to solve the problem (Weak-to-strong CR)
This problem is attributed to the accumulation of pseudo-label errors produced by the model itself, which are exacerbated as self-training progresses.
To mitigate this problem, existing methods distinguish the model prediction processes for supervised outputs and pseudo-labels.
In other words, the supervised outputs and pseudo-labels are obtained from different predictions, respectively.
Some studies \cite{sohn2020fixmatch,zou2020pseudoseg,zhong2021pixel,yang2023revisiting} have adopted the weak-to-strong consistency paradigm from the perspective of input separation, which supervises the prediction from a strongly perturbed input using the pseudo-label generated from its weakly perturbed counterpart (\cref{fig1:framework_comparison}a).

%% Existing Methods to solve the problem (Mean Teacher)
Another method to obtain different prediction views involves using network perturbation based on a teacher-student structure \cite{laine2016temporal,tarvainen2017mean,jin2022semi,wang2022semi,wang2023hunting}.
In particular, Mean Teacher \cite{tarvainen2017mean} is a representative approach in semi-supervised segmentation, where a teacher network is derived using an exponential moving average (EMA) of the student model's weights (\cref{fig1:framework_comparison}b).
Although the Mean Teacher generates somewhat different prediction views between the teacher and student, this method is limited by a coupling problem \cite{ke2019dual}: as training progresses, the teacher and student become tightly linked, and consequently, the teacher's predictions become similar to those of the student.
To mitigate the coupling problem and obtain diverse prediction views, certain studies \cite{liu2022perturbed,na2023switching} have proposed a dual EMA teacher-based framework where two teachers are alternatively updated at every epoch (\cref{fig1:framework_comparison}d).
Among them, PS-MT \cite{liu2022perturbed} produces more reliable pseudo-labels by ensembling the predictions of the two teachers.
By contrast, Na \etal \cite{na2023switching} reported that ensembling predictions can reduce the diversity of pseudo-labels. Thus, they proposed the Dual Teacher framework that alternately activates two teachers in each epoch to provide diversified pseudo-labels.
%
%% Remaining problem 1 (problem of dual teacher frameworks)
Although these studies have demonstrated the benefits of using multiple teachers to mitigate the coupling problem, their ability to provide reliable and diverse pseudo-labels over the course of training remains limited, as they typically focus on only one aspect.
In other words, PS-MT improves the reliability of pseudo-labels but lacks diversity, whereas Dual Teacher suffers from the opposite limitation.
In addition, they have incorporated additional complex components to ensure diversity between the two teachers. In particular, adversarial feature perturbation and a new loss function are used for PS-MT. Further, distinct types of augmentation provided to each teacher (\eg, CutMix or ClassMix) and layer perturbation are used for Dual Teacher.
Consequently, this complexity can hinder the scalability and compatibility of these approaches with existing semi-supervised segmentation methods.

%% Remaining problem 2 (cost problem of co-training method)
Instead of using EMA-based teachers, a co-training paradigm \cite{qiao2018deep,ouali2020semi,fan2022ucc,ke2019dual,chen2021semi,wang2023conflict} has been widely used to expand prediction views (\cref{fig1:framework_comparison}c). This paradigm involves simultaneously training multiple networks with different initializations in a mutual teaching manner, where each network supervises the others using pseudo-labels generated from its predictions.
Building on this concept, a subsequent study \cite{li2023diverse} has demonstrated that increasing the diversity of pseudo-label views from student networks improves their generalization ability. 
Co-training provides diverse pseudo-label guidance with stability and without concerns regarding the coupling problem. However, its scalability remains constrained due to computational complexity and resource demands.

%% Our approach
In this paper, we propose the PrevMatch framework, which efficiently expands pseudo-label views by maximizing the utilization of previous models obtained during training, as depicted in \cref{fig1:framework_comparison}e.
The PrevMatch framework is based on two main ideas.
First, to efficiently address the coupling problem, we revisit the utilization of temporal knowledge. Specifically, we save several models at specific epochs during training and utilize their predictions as additional guidance, referred to as \textit{previous guidance}, which acts as a regularizer in conjunction with standard guidance. 
Second, we design a highly randomized ensemble strategy to maximize the effectiveness of utilizing the previous guidance.
This approach involves selecting a random number of models from those previously saved and ensembling their predictions using randomized weights. 
These strategies can efficiently provide diverse and reliable pseudo-labels while avoiding the complexities inherent in dual EMA- and co-training-based approaches.
Extensive experiments conducted across various evaluation protocols on the Pascal, Cityscapes, and ADE20K datasets reveal that the proposed PrevMatch significantly improves the performance of different baseline methods. In particular, PrevMatch achieves a 0.9-5.4\% mean IoU improvement on Pascal VOC with a 92-label setting.

\section{Related Work}
\subsection{Semi-supervised Semantic Segmentation}
Recent studies have focused on developing improved methods based on weak-to-strong consistency, Mean Teacher, and co-training paradigms.
PseudoSeg \cite{zou2020pseudoseg} adopts a weak-to-strong paradigm based on a single network. CPS \cite{chen2021semi} and GCT \cite{ke2020guided} employ two networks for co-training and demonstrate that co-training outperforms the Mean Teacher approach.
UniMatch \cite{yang2023revisiting} introduces two key enhancements to FixMatch: dual-stream image-level perturbation and feature-level perturbation strategies.
Subsequent studies have integrated the weak-to-strong paradigm (i.e., input perturbation) into the Mean Teacher or co-training approaches (i.e., network perturbation), demonstrating performance improvements.
A stream of research that combines the weak-to-strong paradigm with the Mean Teacher structure has proposed techniques such as advanced data augmentation \cite{hu2021semi,zhao2023augmentation,zhao2023instance}, prototype learning \cite{xu2022semi}, curriculum learning \cite{ma2023enhanced}, symbolic reasoning \cite{liang2023logic}, and dual Mean Teacher \cite{liu2022perturbed, na2023switching}.
Another stream of research, combining the weak-to-strong paradigm with co-training, has developed techniques such as a shared backbone with multiple heads \cite{fan2022ucc}, conservative-progressive learning \cite{fan2023conservative}, and increasing the diversity of co-training \cite{li2023diverse}. 

The proposed PrevMatch framework can be seen as simplifying and extending PS-MT \cite{liu2022perturbed} and Dual Teacher \cite{na2023switching} while also enhancing the efficiency of Diverse Co-training \cite{li2023diverse}.
Specifically, we eliminate the complex components used in PS-MT and Dual Teacher, such as distinct augmentation types for each epoch, layer/adversarial feature perturbations, EMA teachers, and a new loss function.
Instead, PrevMatch reuses the previous models and the weakly perturbed input used in the standard flow, thereby improving the simplicity and compatibility of the overall framework.
Furthermore, we provide reliable and diverse pseudo-labels to the student network through a highly randomized ensemble strategy.
Moreover, although both PrevMatch and Diverse Co-training provide diverse pseudo-label views, PrevMatch operates on a single trainable network with fixed previous models. This facilitates greater efficiency in terms of computational and memory costs.

\subsection{Temporal Knowledge in Semi-supervised Learning}
In the context of semi-supervised learning, several studies have leveraged temporal knowledge obtained from previous training stages.
Temporal Ensembling \cite{laine2016temporal} accumulates predictions for unlabeled samples across different epochs using the EMA approach and enforces consistency between the current and EMA predictions. 
Mean Teacher \cite{tarvainen2017mean} averages the model weights across the training steps using the EMA approach, producing a more stable teacher network. 
PS-MT \cite{liu2022perturbed} and Dual Teacher \cite{na2023switching}, inheriting the spirit of the Mean Teacher method, aim to further exploit temporal knowledge by adopting two EMA teachers.
Moreover, TC-SSL \cite{zhou2020time} and ST++ \cite{yang2022st++} methods implement a filtering mechanism to exclude less-informative unlabeled samples using specific scoring criteria measuring temporal consistency.
In contrast to existing methods, we directly utilize previous models and generate pseudo-label guidance from their predictions without relying on a complex pipeline of dual EMA-based methods or a filtering mechanism.

\section{Method}
This section describes the preliminaries and overall training flow for semi-supervised semantic segmentation.
The PrevMatch method is also introduced here.

\subsection{Preliminaries \& Overall Workflow}
\label{subsec3:preliminary}
Semi-supervised semantic segmentation aims to fully utilize unlabeled images $D_u = \{x^{u}_{i}\}$, given only a limited number of labeled images $D_l = \{(x^{l}_{i}, y^{l}_{i})\}$.
In general semi-supervised learning, the objective function is divided into supervised loss $L_s$ and unsupervised loss $L_u$ as follows: 
\begin{equation}
\label{eq1}
L = \frac{1}{2}(L_s + L_u).
\end{equation}

\begin{enumerate}[label=\arabic*)]
    \item \textbf{Supervised Flow}
    In supervised flow, a segmentation network $f$ receives a labeled image $x^{l}$ and generates the corresponding predicted class distribution $p^{l}$. Subsequently, the supervised loss is calculated using pixel-wise cross-entropy $H$ between the prediction and the ground-truth label. This can be formulated as follows.
    \begin{equation}
      L_s = \sum H(y^l, p^l).
      \label{eq2}
    \end{equation}

    \item \textbf{Unsupervised -- Standard Flow}
    Following the success of the weak-to-strong consistency paradigm popularized by FixMatch \cite{sohn2020fixmatch}, most semi-supervised segmentation methods have adopted this paradigm.
    Concretely, two perturbed images ($x^w$ and $x^s$) are obtained by applying weak and strong augmentations to an unlabeled image $x^u$.
    The network $f$ receives the two perturbed images and outputs the predicted class distributions $p^w$ and $p^s$.
    As there is no ground-truth label for the unlabeled image, a pseudo-label for the weakly perturbed prediction is obtained by: $y^w \text{= argmax}(p^w)$. Then, this pseudo-label is used to supervise the strongly perturbed prediction, as depicted in the standard flow presented in \cref{fig1:framework_comparison}e (red arrow). 
    This consistency term can be formulated as:
    \begin{equation}
    \label{eq3}
    L_{u(standard)} = \sum\mathbb{1}(max(p^w) \geq \tau) H(y^w, p^s),
    \end{equation}
    where $\mathbb{1}(\cdot)$ is an indicator function, and $\tau$ is a confidence threshold used to ignore noise in a pseudo-label.
    
    \item \textbf{Unsupervised -- Proposed Flow}
    To obtain additional pseudo-label guidance, the same weakly perturbed image, $x^w$, used in the standard branch, is fed into the previous model branch. 
    Subsequently, $k$ models, where $k$ is chosen randomly from $\{1,2,..., K\}$, are selected from the saved list of previous models, and $k$ predictions for $x^w$ are generated. Previous guidance, $p^{w'}$, is obtained by aggregating the $k$ predictions using randomized ensemble weights to improve the diversity of the pseudo-label view. As illustrated by the blue arrow in \cref{fig1:framework_comparison}e, the previous guidance functions as additional regularizers, supervising the strongly perturbed prediction. This can be formulated as follows.
    \begin{equation}
    \label{eq5}    
    L_{u(prev)} = \sum\mathbb{1}(max(p^{w'}) \geq \tau) H(y^{w'}, p^s).
    \end{equation}

\end{enumerate}
Finally, the total loss of the unsupervised flow is defined by combining the standard and proposed flow losses as follows:
\begin{equation}
    L_u = L_{u(standard)} + \lambda \cdot L_{u(prev),}
  \label{eq6}
\end{equation}
where $\lambda$ denotes the weight of the proposed flow.

\subsection{Previous Guidance}
\label{subsec3:previous_guidance}
To address the coupling problem between student and teacher networks, we directly utilize previous models and generate pseudo-label guidance from their predictions, instead of using Mean Teacher or co-training approaches.
In particular, we store multiple models at different epochs that meet the specified criteria during training and use their pseudo-labels as additional guidance, referred to as previous guidance.
As training progresses, the decoupling between the student and previously saved teachers increases, allowing for the acquisition of different prediction views. However, as decoupling becomes more pronounced, the positive effects of self-training from correct pseudo-labels may diminish due to the use of outdated teachers. Therefore, previous guidance is used in conjunction with the standard guidance obtained from the current student. In addition, we define the maximum length ($N$) of the list for storing previous models and replace the oldest teacher with a newer one when this limit is exceeded to avoid using excessively outdated teachers.

Formally, one previous model is randomly selected from the previous model list $\{T_1, T_2, ..., T_N\}$.
Then, this model processes the same weakly perturbed image, $x^w$, used in the standard flow and produces predictions $p^{w'}$. The previous guidance is obtained by: $y^{w'} \text{= argmax}(p^{w'})$, and it supervises the strongly perturbed prediction according to Eq. \eqref{eq5}.
In this way, we produce diverse pseudo-labels, as in prior studies \cite{liu2022perturbed, na2023switching, chen2021semi, li2023diverse}, by leveraging different previous models encompassing varied perspectives of temporal knowledge, without complex additional components or heavy computational burden.

\subsubsection{Save Criteria.}
\label{subsubsec:save_criteria}
In this approach, storing the appropriate previous models is crucial for generating diverse and reliable pseudo-labels.
Some studies \cite{huang2017snapshot, wang2022efficient} that employ intermediate models for ensembling in image recognition save the model at regular intervals (\eg, every 20 epochs in a total of 100 epochs).
In contrast to standard image recognition, the weights and performances of the neural network can vary significantly during the optimization process in semi-supervised segmentation (i.e., our case) due to label scarcity and a pronounced class imbalance.
Therefore, we save the model when it achieves the best performance on the validation set to ensure the stability of the previous guidance. In addition, this approach does not require additional hyperparameter searches, such as determining appropriate intervals. Therefore, we adopt our saving approach to achieve stability and simplicity.

The efficacy of the previous guidance can be intuitively explained as follows.
When using only standard guidance, two training scenarios arise based on whether the prediction is correct or incorrect. In cases where the prediction is incorrect, the network is trained in the wrong direction.
In contrast, given the standard and previous guidance, four scenarios can be considered based on whether they are correct or incorrect (standard-previous): (1) correct-correct, (2) correct-incorrect, (3) incorrect-correct, and (4) incorrect-incorrect. 
Through case (3), the network receives an additional opportunity to be guided in the right direction, away from the wrong one
(Refer to \cref{supp_sec:training_stability} in the supplementary material for further discussion).

\subsection{Maximizing Efficacy of Previous Guidance}
\label{subsec3:maximizing}
One way to improve the reliability of the predictions involves using network ensemble techniques.
These techniques have been widely used in various domains as a promising method for improving performance.
In particular, several studies \cite{huang2017snapshot, wang2022efficient} in the field of image recognition have demonstrated that ensembles of multiple intermediate models obtained during training also improve prediction accuracy and diversity.
Therefore, we utilize network ensembling to improve the reliability of the previous guidance.
In designing this method, we also consider computational complexity and pseudo-label diversity.
To this end, given the list that includes $N$ previous models, we randomly select $K$ ($K \leq N$) models for each iteration to mitigate the increase in computational complexity while ensuring pseudo-label diversity.
However, this approach, which always ensembles $K$ models, may not guarantee the diversity of pseudo-labels, as noted in Dual Teacher \cite{na2023switching}. This problem can worsen for large values of $K$.
Therefore, to provide reliable and diverse pseudo-labels to the student network, we propose a highly randomized ensemble strategy comprising the following two ideas:
\begin{itemize}
    \item 
    \textbf{Random Selection:} For each iteration, we randomly select a varying number of teachers, $k$, ranging from 1 to $K$. For example, $k$=1,2, or 3 can be selected for each iteration when $K$=3. With this approach, selecting a large $k$ tends to yield consistent pseudo-labels, enhancing reliability, while a small $k$ contributes to increased diversity in the pseudo-labels. In addition, computational costs are lower than those incurred when using a fixed number $K$. This strategy enables the student network to obtain stable and diverse guidance, which functions as a robust regularizer, aiding network optimization.

    \item 
    \textbf{Random Weights:} Regarding ensemble weights, we propose a random aggregation strategy that averages $k$ predictions using random weights for each iteration instead of a simple average of $k$ predictions. In particular, the $k$ selected teachers receive $x^w$ and output $k$ predictions, $\{p^{w'}_1, p^{w'}_2, ..., p^{w'}_k\}$. 
    The final previous guidance is then obtained by aggregating these predictions using random weights as follows:
    \begin{equation}
    \label{eq7}
    p^{w'} = \sum_{i=1}^{k} w_i \cdot p^{w'}_i,
    \end{equation}
    where $w_i$ is derived from a Dirichlet distribution as follows: $\{w_1, w_2, ..., w_k\}\sim\text{Dir}(\alpha_1, \alpha_2, ..., \alpha_k)$. Note that the sum of $w_i$ is one.
    This approach explores all combinations of previous guidance in a continuous space, expanding the original pseudo-label space beyond a simple average.
\end{itemize}
In this way, the proposed randomized ensemble strategy probabilistically generates pseudo-labels with varying degrees of reliability and diversity across the training process, thereby enabling broader and more flexible exploration of the pseudo-label space.

\section{Experiments}

\subsection{Datasets}
(1) Pascal VOC dataset \cite{everingham2010pascal} contains 10,582 training images, segmented into two subsets based on annotation quality: 1,464 images constitute the high-quality subset, whereas the remaining 9,118 images form the coarse subset.
In particular, two training protocols are considered based on the criteria for selecting labeled images: \textit{Original} protocol whereby labeled images are exclusively sourced from the high-quality subset. \textit{Blended} protocol that entails a random selection of labeled images from the total dataset. (2) Cityscapes dataset \cite{cordts2016cityscapes}, tailored for the semantic analysis of urban street scenes, comprises 2,975 high-resolution images for training and 500 images for validation, primarily focusing on 19 categories within urban environments. (3) ADE20K dataset \cite{zhou2017scene}, containing more diverse scenes and a greater number of classes (150 categories) than the COCO dataset \cite{lin2014microsoft}, comprises 20,210 images for training and 2,000 images for validation. 
In all protocols, training images not selected as labeled images are utilized as unlabeled images.

\subsection{Implementation Details}
We select various methods that potentially exhibit confirmation bias and coupling problems as baselines, which use the input perturbation method based on a single network.
For the four baseline methods (FixMatch, UniMatch, AugSeg, and MPMC), we employ ResNet-50/101 \cite{he2016deep} as the encoder and DeepLabV3+ \cite{chen2018encoder} as the segmentation head. For UniMatchV2, we use the DINOv2 \cite{oquab2023dinov2} encoder and the DPT \cite{ranftl2021vision} segmentation model.
Specifically, UniMatch and UniMatchV2 are the primary baselines in our study. For these two baselines, each mini-batch comprises 8 labeled and 8 unlabeled images. Furthermore, they are trained for 80, 240/180, and 40/60 epochs for the PASCAL, Cityscapes, and ADE20K datasets, respectively.
The learning rates are initially set to 1e-3/5e-6, 5e-3/5e-6, and 4e-3/5e-6 for these datasets, respectively, and are managed using a polynomial learning rate scheduler. Moreover, we use 321/518, 801/798, and 513/518 random crops for these datasets, respectively.
For image augmentation, we use common weak (\eg, resize, crop, and flip) and strong (\eg, color transformations, grayscale, cutmix, and blur) data augmentations, as in UniMatch \cite{yang2023revisiting}.
For the hyperparameters of standard consistency flow, $\tau$ is set to 0 for Cityscapes and 0.95 for the other datasets.

For the hyperparameters of PrevMatch, $\tau$ is set to 0.9 for Pascal, 0 for Cityscapes, and 0.95 for ADE20k. The maximum length ($N$) of the previous list is set to eight for Pascal and Cityscapes and six for ADE20K. The upper bound number $K$ for the random selection is set to three. For weight $\lambda$, as previous guidance should correct the model before it becomes overfitted in the wrong direction (i.e., confirmation bias \cite{arazo2020pseudo}) during the middle of training, setting an appropriate weight $\lambda$ is crucial.
Additionally, as the initial model typically exhibits poor performance, we implement a warmup schedule for the weight of the proposed flow, similar to the commonly used learning rate decay scheduling that includes a warmup phase. Refer to \cref{supp_sec:loss_weight} in the supplementary material for further details.

% Original table
\begin{table*}[t]
\centering
\resizebox{0.97\linewidth}{!}{%
\begin{tabular}{llc|ccccc}
\toprule
\textbf{Pascal \small[Original set]} & & Encoder & 92 & 183 & 366 & 732 & 1464 \\
\midrule
AugSeg \cite{zhao2023augmentation} & \scriptsize{\textcolor{gray}{[CVPR'23]}} & RN-50 & 64.2 & 72.2 & 76.2 & 77.4 & 78.8 \\
Dual Teacher \cite{na2023switching} & \scriptsize{\textcolor{gray}{[NeurIPS'23]}} & RN-50 & 70.8 & 74.5 &	76.4 & 77.7 & 78.2 \\
\midrule
FixMatch \cite{sohn2020fixmatch} & \scriptsize{\textcolor{gray}{[NeurIPS'20]}} & RN-50 & 63.8 & 70.3 & 73.2 & 76.8 & 77.8\\
 + \textbf{PrevMatch} & & RN-50 & \textbf{66.1} \green{(\small{+2.3})} & \textbf{73.1} \green{(\small{+2.8})} & \textbf{75.7} \green{(\small{+2.5})} & \textbf{77.8} \green{(\small{+1.0})}& \textbf{78.5} \green{(\small{+0.7})}\\
\midrule
UniMatch \cite{yang2023revisiting} & \scriptsize{\textcolor{gray}{[CVPR'23]}} & RN-50 & 71.9 & 72.5 & 76.0 & 77.4 & 78.7 \\
 + \textbf{PrevMatch} & & RN-50 & \textbf{73.4} \green{(\small{+1.5})} & \textbf{75.4} \green{(\small{+2.9})} & \textbf{77.5} \green{(\small{+1.5})} & \textbf{78.6} \green{(\small{+1.2})}& \textbf{79.3} \green{(\small{+0.6})}\\
\midrule
\midrule
U$^{2}$PL \cite{wang2022semi} & \scriptsize{\textcolor{gray}{[CVPR'22]}} & RN-101 &	68.0 & 69.2 & 73.7 & 76.2 &	79.5 \\
PS-MT \cite{liu2022perturbed} & \scriptsize{\textcolor{gray}{[CVPR'22]}} & RN-101	& 65.8 &	69.6 & 76.6 & 78.4 & 80.0 \\
Diverse Co-T.(3cps) \cite{li2023diverse} & \scriptsize{\textcolor{gray}{[ICCV'23]}} & RN-101 & 75.4 & 76.8 & 79.6 & 80.4 &	81.6 \\
CorrMatch \cite{sun2024corrmatch} & \scriptsize{\textcolor{gray}{[CVPR'24]}} & RN-101 & 76.4 & 78.5 & 79.4 & 80.6 & 81.8 \\
U$^{2}$PL+ \cite{wang2024using} & \scriptsize{\textcolor{gray}{[IJCV'24]}} & RN-101 & 69.3 & 73.4 & 75.0 & 77.1 & 79.5 \\
\midrule
AugSeg \cite{zhao2023augmentation} & \scriptsize{\textcolor{gray}{[CVPR'23]}} & RN-101 & 71.1 & 75.5 & 78.8 & 80.3 & 81.4 \\
 + \textbf{PrevMatch} & & RN-101 & \textbf{73.4} \green{(\small{+2.3})} & \textbf{77.3} \green{(\small{+1.8})} & \textbf{80.0} \green{(\small{+1.2})} & \textbf{80.9} \green{(\small{+0.6})} & \textbf{82.0} \green{(\small{+0.6})} \\
\midrule
UniMatch \cite{yang2023revisiting} & \scriptsize{\textcolor{gray}{[CVPR'23]}} & RN-101	& 75.2 & 77.2 &	78.8 &	79.9 &	81.2 \\
 + \textbf{PrevMatch} & & RN-101 & \textbf{77.0} \green{(\small{+1.8})} & \textbf{78.5} \green{(\small{+1.3})} & \textbf{79.6} \green{(\small{+0.8})} & \textbf{80.4} \green{(\small{+0.5})} & \textbf{81.6} \green{(\small{+0.4})} \\
\midrule
\midrule
AllSpark \cite{wang2024allspark} & \scriptsize{\textcolor{gray}{[CVPR'24]}} & MiT-B5 & 76.1 & 78.4 & 79.8 & 80.8 & 82.1\\
SemiVL \cite{hoyer2023semivl} & \scriptsize{\textcolor{gray}{[ECCV'24]}} & CLIP-B & 84.0 & 85.6 & 86.0 & 86.7 & 87.3\\
\midrule
UniMatchV2-S \cite{yang2025unimatch} & \scriptsize{\textcolor{gray}{[TPAMI'25]}} & DINOv2-S & 79.0 & 85.5 & 85.9 & 86.7 & 87.8 \\
 + \textbf{PrevMatch} & & DINOv2-S & \textbf{84.4} \green{(\small{+5.4})} & \textbf{86.2} \green{(\small{+0.7})} & \textbf{86.7} \green{(\small{+0.8})} & \textbf{87.5} \green{(\small{+0.8})} & \textbf{88.2} \green{(\small{+0.4})} \\
\midrule
UniMatchV2-B \cite{yang2025unimatch} & \scriptsize{\textcolor{gray}{[TPAMI'25]}} & DINOv2-B & 86.3 & 87.9 & 88.9 & 90.0 & 90.8 \\
 + \textbf{PrevMatch} & & DINOv2-B & \textbf{87.2} \green{(\small{+0.9})} & \textbf{88.6} \green{(\small{+0.7})} & \textbf{90.5} \green{(\small{+1.6})} & \textbf{90.6} \green{(\small{+0.6})} & \textbf{91.3} \green{(\small{+0.5})}\\
\bottomrule
\end{tabular}
}
\caption{Comparison with state-of-the-art methods on the \textit{Original} protocol of Pascal VOC. The evaluation metric is the mean IoU (\%).}
\label{table:original}
\end{table*}

% Blended table
\begin{table}[t!]
\centering
\resizebox{0.99\linewidth}{!}{%
\begin{tabular}{lc|ccc}
\toprule
\textbf{Pascal \small[Blended set]} & Encoder & 1/16 & 1/8 & 1/4 \\
\midrule
Mean Teacher \cite{tarvainen2017mean} & RN-50 & 66.8 & 70.8 & 73.2\\
PS-MT \cite{liu2022perturbed} & RN-50 & 72.8 & 75.7 & 76.4 \\
CorrMatch \cite{sun2024corrmatch} & RN-101 & 77.6 & 77.8 & 78.3 \\
AllSpark \cite{wang2024allspark} & MiT-B5 & 78.3 & 80.0 & 80.4 \\
\midrule
FixMatch \cite{sohn2020fixmatch} & RN-50 & 72.5 & 73.0 & 74.2\\
 + \textbf{PrevMatch} & RN-50 & \textbf{73.9} & \textbf{74.3} & \textbf{75.1}\\
\;\;\;Gain ($\Delta$) & & \green{(\small{+1.4})} & \green{(\small{+1.3})} & \green{(\small{+0.9})}\\
\midrule
UniMatch \cite{yang2023revisiting} & RN-101 & 76.5 &  77.0 & 77.2 \\ % 74.5 & 75.8 & 76.1 \\
 + \textbf{PrevMatch} & RN-101 & \textbf{77.8} & \textbf{78.0} & \textbf{78.0} \\
\;\;\;Gain ($\Delta$) & & \green{(\small{+1.3})} & \green{(\small{+1.0})} & \green{(\small{+0.8})}\\
\midrule
UniMatchV2-S \cite{yang2025unimatch} & DINOv2-S & 82.2 & 83.6 & 83.8 \\
 + \textbf{PrevMatch} & DINOv2-S & \textbf{83.3} & \textbf{84.5} & \textbf{84.5} \\
\;\;\;Gain ($\Delta$) & & \green{(\small{+1.1})} & \green{(\small{+0.9})} & \green{(\small{+0.7})}\\
\bottomrule
\end{tabular}
}
\caption{Comparison with state-of-the-art methods on the \textit{Blended} protocol of Pascal VOC dataset.}
\label{table:blend}
\end{table}

\subsection{Comparison with State-of-the-Art Methods}
\noindent\textbf{PASCAL VOC. } Two main experiments are conducted on the \textit{Original} and \textit{Blended} protocols.
In \cref{table:original}, applying PrevMatch to each baseline consistently improves performance across all baseline methods. Notably, we observe that PrevMatch yields the most improvements in settings with fewer labels (92 and 183). In the 92-label setting (ResNet-101), it improves AugSeg, UniMatch, and MPMC by 2.3\%, 1.8\%, and 1.2\%, respectively. 
Furthermore, even on UniMatchV2, which already exhibits strong performance with its transformer structure, PrevMatch further enhances performance.
Moreover, \cref{table:blend} reports results on the \textit{Blended} protocol. PrevMatch consistently improves the performance of all baseline methods.

\noindent\textbf{Cityscapes.}
Experimental results for the Cityscapes validation set are presented in \cref{table:cityscapes}.
The results indicate that applying PrevMatch to the four baseline approaches consistently improves their performance across all label partitions. Similar to the results reported in \cref{table:original}, PrevMatch significantly improves performance in setups with fewer labels. In particular, in the 1/16 label setup, PrevMatch improves AugSeg, UniMatch, UniMatchV2-S, and UniMatchV2-B by 1.4\%, 1.1\%, 1.1\%, and 0.8\%, respectively.

% Double column Cityscapes table
\begin{table}[t]
\centering
\resizebox{1.0\linewidth}{!}{%
\begin{tabular}{l|cccc}
\toprule
\multirow{2}{*}{\textbf{Cityscapes}} & 1/16 & 1/8 & 1/4 & 1/2 \\
& (186) & (372) & (744) & (1488) \\
\midrule
\multicolumn{5}{c}{ResNet-101}\\
\midrule
PS-MT \cite{liu2022perturbed} & - & 76.9 & 77.6 & 79.1 \\
Diverse Co-T.(3cps) \cite{li2023diverse} & 75.7 & 77.4 & 78.5 & - \\
Dual Teacher \cite{na2023switching} & 76.8 & 78.4 & 79.5 & 80.5 \\
CorrMatch \cite{sun2024corrmatch} & 77.3 & 78.5 & 79.4 & 80.4 \\
\midrule
AugSeg \cite{zhao2023augmentation} & 75.2 & 77.8 & 79.6 & 80.4 \\
 + \textbf{PrevMatch} & \textbf{76.6} & \textbf{78.9} & \textbf{80.3} & \textbf{80.8}\\
\;\;\;Gain ($\Delta$) & \green{(\small{+1.4})} & \green{(\small{+1.1})} & \green{(\small{+0.7})} & \green{(\small{+0.4})}\\
\midrule
UniMatch \cite{yang2023revisiting} & 76.6 & 77.9 & 79.2 & 79.5 \\
 + \textbf{PrevMatch} & \textbf{77.7} & \textbf{78.9} & \textbf{80.1} & \textbf{80.1} \\
\;\;\;Gain ($\Delta$) & \green{(\small{+1.1})} & \green{(\small{+1.0})} & \green{(\small{+0.9})} & \green{(\small{+0.6})}\\
\midrule
\midrule
\multicolumn{5}{c}{Tansformer-based}\\
\midrule
AllSpark \cite{wang2024allspark} & 78.3 & 79.2 & 80.6 & 81.4 \\
SemiVL \cite{hoyer2023semivl} & 77.9 & 79.4 & 80.3 & 80.6\\
\midrule
UniMatchV2-S \cite{yang2025unimatch} & 80.6 & 81.9 & 82.4 & 82.6\\
 + \textbf{PrevMatch} & \textbf{81.7} & \textbf{82.7} & \textbf{83.2} & \textbf{83.2}\\
\;\;\;Gain ($\Delta$) & \green{(\small{+1.1})} & \green{(\small{+0.8})} & \green{(\small{+0.8})} & \green{(\small{+0.6})}\\
\midrule
UniMatchV2-B$^\dagger$ \cite{yang2025unimatch} & 82.6 & 83.3 & 83.5 & 83.9 \\
 + \textbf{PrevMatch$^\dagger$} & \textbf{83.4} & \textbf{84.1} & \textbf{84.5} & \textbf{84.8}\\
\;\;\;Gain ($\Delta$) & \green{(\small{+0.8})} & \green{(\small{+0.8})} & \green{(\small{+1.0})} & \green{(\small{+0.9})}\\
\bottomrule
\end{tabular}
}
\caption{Comparison with state-of-the-art methods on the Cityscapes dataset. $\dagger$: The results of UniMatchV2 (DINOv2-B) were reproduced with a training resolution of 756$^2$ due to limitations in our available memory.}
\label{table:cityscapes}
\end{table}

\noindent\textbf{ADE20K. }
\cref{table:ade20k} lists results for the large-scale and challenging dataset ADE20K. PrevMatch consistently improves baseline performance across all partitions, suggesting that PrevMatch is also effective on the large-scale dataset.

% Double column ADE20K table
\begin{table}[t]
\centering
\resizebox{1.0\linewidth}{!}{%
\begin{tabular}{lccccc}
\toprule
\textbf{ADE20K} & 1/128 & 1/64 & 1/32 & 1/16 & 1/8 \\
\midrule
UniMatch-RN50 \cite{yang2023revisiting} & 13.6 & 18.3 & 23.9 & 27.2 & 30.9\\
 + \textbf{PrevMatch} & \textbf{15.4} & \textbf{19.6} & \textbf{24.9} & \textbf{28.3} & \textbf{31.6}\\
\;\;\;Gain ($\Delta$) & \green{(\small{+1.8})} & \green{(\small{+1.3})} & \green{(\small{+1.0})} & \green{(\small{+1.1})} & \green{(\small{+0.7})}\\
\midrule
UniMatchV2-S \cite{yang2025unimatch} & 20.9 & 29.1 & 34.5 & 37.0 & 39.4\\
 + \textbf{PrevMatch} & \textbf{22.2} & \textbf{29.8} & \textbf{35.2} & \textbf{37.8} & \textbf{40.0}\\
\;\;\;Gain ($\Delta$) & \green{(\small{+1.3})} & \green{(\small{+0.7})} & \green{(\small{+0.7})} & \green{(\small{+0.8})} & \green{(\small{+0.6})}\\
\midrule
UniMatchV2-B \cite{yang2025unimatch} & 25.0 & 33.7 & 40.5 & 42.4 & 45.1 \\
 + \textbf{PrevMatch} & \textbf{25.7} & \textbf{35.1} & \textbf{41.4} & \textbf{43.4} & \textbf{46.0} \\
\;\;\;Gain ($\Delta$) & \green{(\small{+0.7})} & \green{(\small{+1.4})} & \green{(\small{+0.9})} & \green{(\small{+1.0})} & \green{(\small{+0.9})}\\
\bottomrule
\end{tabular}
}
\caption{Evaluation results on the large-scale dataset ADE20K, as reproduced in our experimental setup.}
\label{table:ade20k}
\end{table}

% Ablation on components
\begin{table}[t]
\centering
\resizebox{1.0\linewidth}{!}{%
\begin{tabular}{cccc|cc}
\toprule
Previous & Simple & Random  & Random & w/ UniM. & w/ UniM.V2\\
Guidance & Ensemble & Selection & Weights & 92 & 92 \\
\midrule
- & - & - & - & 71.9 & 79.0 \\
\checkmark & - & - & - & 72.7 & 82.2 \\
\checkmark & \checkmark & - & - & 72.7 & 82.7 \\
\checkmark & - & \checkmark & - & 73.2 & 83.8 \\
\midrule
\checkmark & - & \checkmark & \checkmark & \textbf{73.4} & \textbf{84.4}\\
\bottomrule
\end{tabular}
}
\caption{Ablation study of PrevMatch's components based on UniMatch (ResNet-50) and UniMatch V2 (DINOv2-S) on Pascal VOC with 92 labels. For the ensemble, we set $K$=3.}
\label{table:ablation_components}
\end{table}

\section{Discussion}
\subsection{Ablation Study}
The effects of the individual components are investigated, and the results are presented in \cref{table:ablation_components}.
The first row indicates the baselines. All components of the proposed method consistently achieve performance gains.
In particular, previous guidance, which randomly selects one previous model from the previous list for each iteration to generate additional guidance, surpasses the baselines by 0.8\% and 3.2\% in the 92-label settings, respectively.
Regarding the number of models ($k$) for the network ensemble, we experiment with simple ensemble (fixed $k$) or random selection (random $k$) strategies.
The result (third row) obtained using fixed $k$ exhibits minor performance gains. In contrast, the results using random $k$ (fourth row) indicate significantly improved performance.
This implies that a fixed $k$ ensemble improves the reliability of pseudo-labels but limits the diversity, as noted in Dual Teacher \cite{na2023switching}, whereas a random $k$ strategy can provide reliable and diverse guidance to the model. 
In addition, using random weights for network ensembling contributes to performance gains, indicating that it expands the original pseudo-label space by generating diversified guidance.
Refer to \cref{supp_table:ablation_components,supp_table:ablation_max_len,supp_table:ablation_save_criteria,supp_table:ablation_upper_K} in the supplementary material for additional ablation studies on different protocols and the hyperparameters of each component.

\begin{table}[t]
\resizebox{1.0\linewidth}{!}{%
\centering
\begin{tabular}{rccccccccc}
& \rotatebox{80}{\textbf{chair}} & \rotatebox{80}{\textbf{sofa}} & \rotatebox{80}{\textbf{car}} & \rotatebox{80}{\textbf{train}} & \rotatebox{80}{\textbf{m.bike}} & \rotatebox{80}{\textbf{bicycle}} & \rotatebox{80}{\textbf{d.table}} & \rotatebox{80}{\textbf{backgr.}} \\
\midrule
UniMatch & 8 & 33 & 80 & 80 & 75 & 59 & 55 & 91 \\
+ PrevMatch & 21 & 46 & 84 & 84 & 77 & 61 & 57 & 93 \\
Gain ($\Delta$) & \cellcolor{green!30}13 & \cellcolor{green!30}13 & \cellcolor{green!15}4 & \cellcolor{green!15}4 & \cellcolor{green!10}2 & \cellcolor{green!10}2 & \cellcolor{green!10}2 & \cellcolor{green!10}2 \\
\bottomrule
\end{tabular}
}
\caption{Class-wise IoU scores (Top 8 classes based on gain) for Pascal VOC with a 92-label partition using ResNet-50.}
\label{table:class_wise_iou_pascal}
\end{table}

\subsection{Analysis of Performance and Training Stability}
\begin{figure}[t]
\centering
\includegraphics[width=\linewidth]{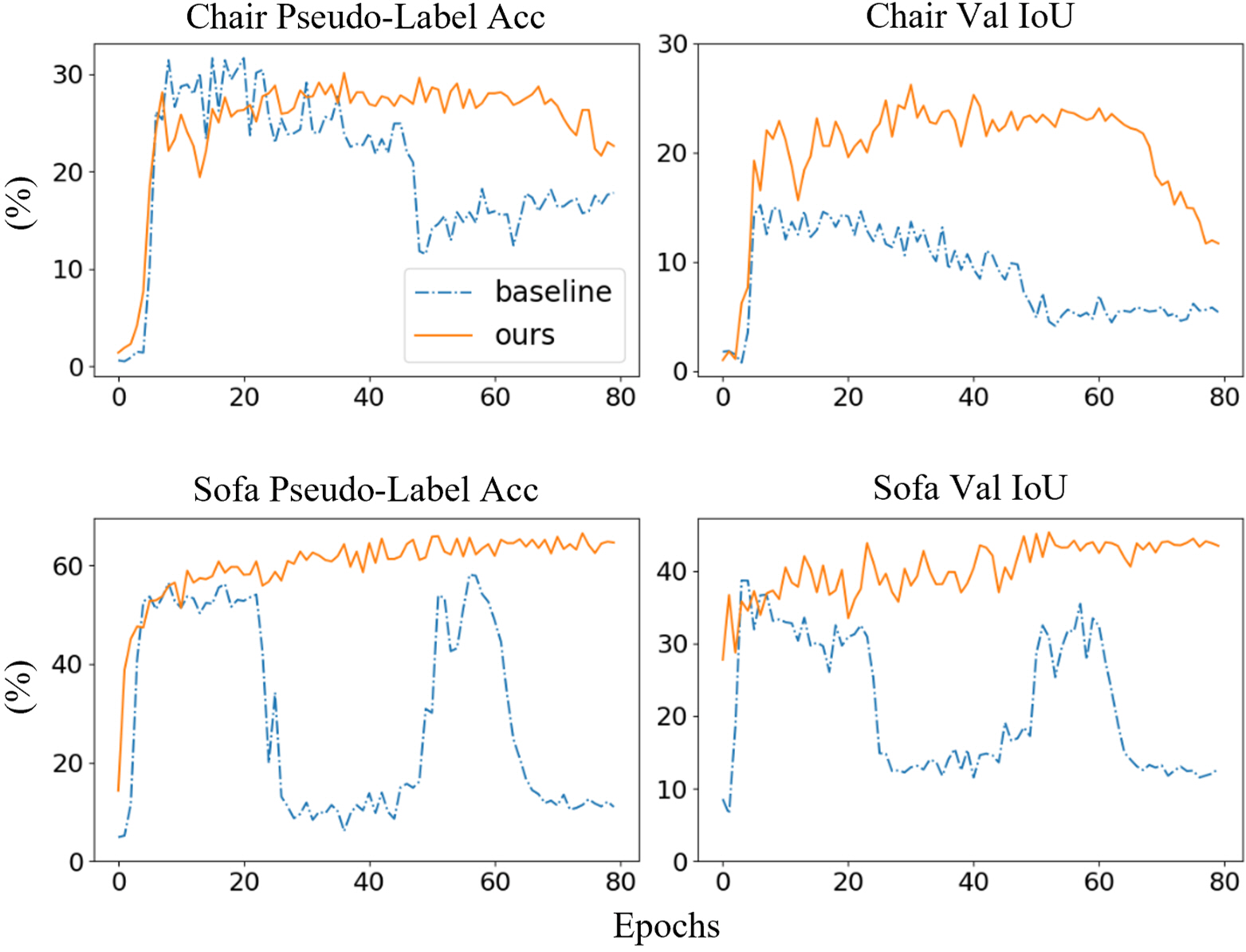}
\caption{Training curves for the chair and sofa classes, illustrating variations in pseudo-label pixel accuracy and validation IoU scores (on Pascal VOC 92-label partition).}
\label{fig:minor_classes_curve}
\end{figure}

\cref{table:class_wise_iou_pascal} lists the category-wise IoU scores and shows that the proposed method achieves notable performance gains for the chair and sofa classes, which were particularly challenging for the UniMatch baseline.
To further explore these poorly behaved classes, we visualize changes in terms of pseudo-label accuracy and validation IoU scores during training, as depicted in \cref{fig:minor_classes_curve}.
In the first row (chair class), the training curve of the baseline pseudo-label accuracy exhibits significant fluctuations, particularly showing a sudden sharp performance drop at approximately 50 epochs. Although the pseudo-label accuracy recovers slightly thereafter, the validation score does not.
In the second row (sofa class), the training curve of the baseline exhibits more severe fluctuations and sharper and more drastic performance drops than in the first row.
This phenomenon can be attributed to catastrophic forgetting where previously learned knowledge is forgotten when acquiring new knowledge \cite{mccloskey1989catastrophic, kirkpatrick2017overcoming}. In other words, self-training and class-imbalance scenarios (i.e., in our scenario) lead to instability in pseudo-label predictions, particularly for poorly behaved classes, due to the lack of labeled data.
In contrast to the baseline, the proposed method shows a smoother training curve without significant fluctuations in either category. This indicates that the proposed training procedure aids in achieving stable optimization for poorly behaved classes that suffer from the forgetting problem (Refer to \cref{supp_sec:additional_analysis} in the supplementary material for additional analysis).

\begin{table}[t!]
\centering
\resizebox{0.99\linewidth}{!}{%
\begin{tabular}{lc|cc}
\toprule
Method & Encoder & \multicolumn{2}{c}{Pascal VOC $\rightarrow$ COCO} \\
\midrule
UniMatch \cite{yang2023revisiting} & RN-50 & 71.9 & 35.6 \\
 + PrevMatch & RN-50 & 73.4 \textbf{\green{(\small{+1.5})}} & 38.5 \textbf{\green{(\small{+2.9})}} \\
\midrule
UniMatchV2-B \cite{yang2023revisiting} & DINOv2-B & 86.3 & 52.9 \\
 + PrevMatch & DINOv2-B & 87.2 \textbf{\green{(\small{+0.9})}} & 55.1 \textbf{\green{(\small{+2.2})}} \\
\bottomrule
\end{tabular}
}
\caption{Evaluation results on seen (Pascal VOC) and unseen (COCO) datasets. COCO contains 80 object classes, 20 of which correspond to the categories in Pascal VOC. The remaining 60 classes were treated as a background class for evaluation. 
The results on COCO were obtained without re-training, using the weights of the model trained on Pascal VOC with a 92-label setting.}
\label{table:generalization}
\end{table}
\subsection{Training Stability and Generalization}
Several studies \cite{li2018visualizing,foret2020sharpness,chen2021vision,shin2023metricgan} across various domains have demonstrated that stable optimization during training results in a flatter loss landscape, thereby improving generalization performance.
Therefore, to assess whether the proposed method's training stability contributes to improved generalization, we conduct additional experiments. \cref{table:generalization} presents evaluation results for models trained on Pascal VOC \cite{everingham2010pascal} and tested on the COCO dataset \cite{lin2014microsoft} without re-training. The results show a larger difference in the score on the unseen dataset (COCO) as compared to that on the seen dataset (Pascal VOC), indicating that PrevMatch improves the model's generalization ability.

\begin{table}[t]
\resizebox{1.0\linewidth}{!}{%
\centering
\begin{tabular}{l|cccccc}
\toprule
\multirow{2.5}{*}{\textbf{Method}} & \multirow{2.5}{*}{\parbox{1.5cm}{\#Trainable \\Networks}} & \multirow{2.5}{*}{\parbox{1.35cm}{Training \\ Time (m)}} & \multirow{2.5}{*}{\parbox{1.8cm}{GPU Mem.\\ Usage (G)}} & \multicolumn{3}{c}{Pascal}\\ \cmidrule{5-7}
& & & & 92 & 183 & 366\\
\midrule
PS-MT & 1 & 24.9 & 28.6 & 65.8 & 69.6 & 76.6\\
Diverse Co-T. & 3 & 22.1 & 40.5 & 75.4 & 76.8 & \textbf{79.6}\\
UniMatch & 1 & 7.8 & 21.9 & 75.2 & 77.2 & 78.8\\
+ PrevMatch & 1 & 9.1 & 22.7 & \textbf{77.0} & \textbf{78.5} & \textbf{79.6}\\
\bottomrule
\end{tabular}
}
\caption{Comparison of time spent using existing methods with a ResNet-101 encoder. Training time per epoch and GPU memory usage were measured using the same environment (two A6000 GPUs) and hyperparameters. We used the open-sourced code provided by the authors.}
\label{table:ablation_efficiency}
\end{table}
\subsection{Efficiency Evaluation}
To evaluate the efficiency of PrevMatch, we compare its training time per epoch and GPU memory usage with existing methods. 
Since UniMatch \cite{yang2023revisiting} relies solely on input separation, it may suffer from instability due to the inherent coupling problem. In contrast, PS-MT \cite{liu2022perturbed}, Diverse Co-training \cite{li2023diverse}, and our proposed PrevMatch address this limitation by integrating both network and input separation techniques. 
In \cref{table:ablation_efficiency}, PrevMatch improves UniMatch by 1.8\% (92-label setting), with a slight cost increase. 
Moreover, PrevMatch outperforms Diverse Co-training, while achieving 2.4 times faster training. Meanwhile, PS-MT, a dual Mean Teacher-based method, requires training time comparable to that of Diverse Co-training due to its complex components, despite being a single trainable network. These results suggest that PrevMatch can efficiently mitigate the coupling problem. Therefore, owing to its computational efficiency and the simplicity of its pipeline reusing the inputs and networks, the proposed method can be easily integrated into any semi-supervised learning method.

\section{Conclusion}
In this work, we introduced the PrevMatch framework, which leverages temporal knowledge obtained during training to efficiently address the issues of tight coupling and confirmation bias that impede stable semi-supervised learning.
The main contributions of PrevMatch include revisiting the use of temporal knowledge and maximizing its effectiveness. 
Experiments were conducted on three benchmark datasets, revealing that incorporating PrevMatch into existing methods significantly improves their performance.
In addition, our findings indicate that leveraging temporal knowledge facilitates stable optimization, particularly for classes that exhibit poor performance and fluctuations.
Finally, the computational efficiency and compatibility of the proposed method facilitate its seamless integration into recent semi-supervised semantic segmentation methods.

\noindent\textbf{Acknowledgements.} \; This research was supported by the BK21 FOUR funded by the Ministry of Education of Korea and National Research Foundation of Korea.

{
    \small
    \bibliographystyle{ieeenat_fullname}
    \bibliography{main}
}

\clearpage
\setcounter{page}{1}
\maketitlesupplementary

\section{Additional Experiments and Ablation Studies}
In this study, experiments were conducted in the following environments: Ubuntu 20.04, Python 3.10.4, PyTorch 1.12.1, and NVIDIA 3090Ti or A6000 GPUs. Unless otherwise specified, subsequent experiments were primarily conducted using the UniMatch baseline with ResNet-50.

\subsection{Comparison with State-of-the-Art on the Priority Protocol of Pascal VOC}
As described in the main paper, the Pascal VOC dataset contains a high-quality annotation subset (1,464 images) and a coarse-quality annotation subset (9,118 images). In addition to the \textit{Original} and \textit{Blended} protocols presented in the main manuscript, the \textit{Priority} protocol has been used in several studies for comparison. The protocol where the selection of labeled images is first derived from the high-quality subset; if not sufficient, it is complemented by additional images from the coarse subset. \cref{supp_table:prior} shows that the proposed method consistently improves baseline methods.

\subsection{Additional Ablation Studies}
Due to space constraints, the ablation study (\cref{table:ablation_components}) in the main body of the paper includes only the 92-label protocol. To further validate the proposed method, we provide additional experiments on the 183- to 1464-label protocols.
\cref{supp_table:ablation_components} shows that the components of PrevMatch consistently improve performance across various evaluation protocols.

\subsection{Previous List Length}
We investigate the effect of the length ($N$) of the previous list that stores the temporal models. 
In \cref{supp_table:ablation_max_len}, the results for $N=1$ and $N=2$ are comparable to those of the baseline.
This suggests that the aforementioned coupling problem may persist because the previous models in the list are continually updated with the latest model when $N$ is small.
The cases of $N=$ 4, 8, and 12 consistently outperform the baseline, revealing that the proposed method is not highly sensitive to hyperparameters.
However, the performance for $N=20$ increases only marginally due to the use of outdated teachers.
Based on this result, we recommend setting the value of $N$ to approximately 5--15\% of the total training epochs, which proves to be appropriate for different datasets (e.g., Pascal=6--10, Cityscapes=8--16, and ADE20k=4--6).

% Blended Priority table
\begin{table}[t!]
\centering
\resizebox{0.99\linewidth}{!}{%
\begin{tabular}{lc|ccc}
\toprule
\textbf{Pascal \small[Priority set]} & Encoder & 1/16 & 1/8 & 1/4 \\
\midrule
U$^{2}$PL \cite{wang2022semi} & RN-101 & 77.2 & 79.0 & 79.3 \\
U$^{2}$PL+ \cite{wang2024using} & RN-101 & 77.2 & 79.4 & 80.2 \\
Dual Teacher \cite{na2023switching} & RN-101 & 80.1 & 81.5 & 80.5 \\
CorrMatch \cite{sun2024corrmatch} & RN-101 & 81.3 & 81.9 & 80.9 \\
AllSpark \cite{wang2024allspark} & MiT-B5 & 81.6 & 82.0 & 80.9 \\
\midrule
FixMatch \cite{sohn2020fixmatch} & RN-50 & 75.4 & 76.5 & 76.9\\
 + \textbf{PrevMatch} & RN-50 & \textbf{76.9} & \textbf{77.6} & \textbf{77.8}\\
\;\;\;Gain ($\Delta$) & & \green{(\small{+1.5})} & \green{(\small{+1.1})} & \green{(\small{+0.9})}\\
\midrule
UniMatch \cite{yang2023revisiting} & RN-101 & 80.9 & 81.9 & 80.4 \\
 + \textbf{PrevMatch} & RN-101 & \textbf{81.4} & \textbf{81.9} & \textbf{80.8} \\
\;\;\;Gain ($\Delta$) & & \green{(\small{+0.5})} & \green{(\small{+0.0})} & \green{(\small{+0.4})}\\
\midrule
UniMatchV2-S \cite{yang2025unimatch} & DINOv2-S & 86.6 & 87.3 & 85.9\\
 + \textbf{PrevMatch} & DINOv2-S & \textbf{87.3} & \textbf{87.8} & \textbf{86.4}\\
\;\;\;Gain ($\Delta$) & & \green{(\small{+0.7})} & \green{(\small{+0.5})} & \green{(\small{+0.5})}\\
\bottomrule
\end{tabular}
}
\caption{Comparison with state-of-the-art methods on the \textit{Priority} protocol of Pascal VOC dataset.}
\label{supp_table:prior}
\end{table}

% Additional ablation on components
\begin{table}[t]
\centering
\resizebox{0.99\linewidth}{!}{%
\begin{tabular}{ccc|ccccc}
\toprule
Previous & Random  & Random & \multicolumn{5}{c}{PASCAL}\\
Guidance & Selection & Weights & 92 & 183 & 366 & 732 & 1464\\
\midrule
- & - & - & 71.9 & 72.5 & 76.0 & 77.4 & 78.7 \\
\checkmark & - & - & 72.7 & 73.8 & 76.6 & 78.1 & 79.0 \\
\checkmark & \checkmark & - & 73.2 & 74.9 & 77.4 & 78.3 & 79.2 \\
\midrule
\checkmark & \checkmark & \checkmark & \textbf{73.4} & \textbf{75.4} & \textbf{77.5} & \textbf{78.6} & \textbf{79.3}\\
\bottomrule
\end{tabular}
}
\caption{Additional ablation studies across different evaluation protocols (UniMatch with ResNet-50).}
\label{supp_table:ablation_components}
\end{table}

% Ablation on list_max_len
\begin{table}[t]
\centering
\resizebox{0.99\linewidth}{!}{%
\begin{tabular}{c|ccccccc}
\toprule
$N$ & Base. & 1 & 2 & 4 & 8 & 12 & 20\\
\midrule
Pascal$_{92}$ & 71.9 & 71.7 & 71.7 & 72.4 & \textbf{72.7} & 72.5 & 71.9 \\
Pascal$_{183}$ & 72.5 & 72.7 & 72.9 & 73.4 & \textbf{73.8} & 73.5 & 73.3 \\
\bottomrule
\end{tabular}
}
\caption{Ablation study for the maximum length ($N$) of the previous list. In this setting, only previous guidance is used (i.e., $K$=1, without ensemble).}
\label{supp_table:ablation_max_len}
\end{table}

\subsection{Upper Bound Number for Random Selection}
To generate reliable and diverse pseudo-labels, we proposed a strategy that randomly selects $k$ models (ranging from 1 to $K$) for each iteration. In this strategy, we explore the performance changes regarding the upper bound number $K$.
\cref{supp_table:ablation_upper_K} indicates that including the ensembling cases ($K>1$, i.e., $k$=1 or $k>$1 are randomly selected) improves the performance significantly compared to the case of $K=1$ (i.e., without ensemble).
In addition, we observe the best results at $K=3$ and a slight performance drop in settings with $K$ greater than 3.
Even for large $K$, a varying number ($k$) of models is selected; however, the proportion of large $k$ values increases with $K$.
This ensures consistent pseudo-labels but reduces their diversity, potentially degrading performance, as mentioned in Dual Teacher.
In conclusion, we select $K=3$ because it adequately satisfies the diversity and reliability requirements of the pseudo-labels.

% Ablation on K
\begin{table}[t!]
\centering
\resizebox{0.99\linewidth}{!}{%
\begin{tabular}{c|ccccc}
\toprule
Upper Bound Number ($K$) & 1 & 2 & 3 & 4 & 5\\
\midrule
Pascal$_{92}$ & 72.7 & 73.1 & \textbf{73.4} & \textbf{73.4} & 73.1 \\
Pascal$_{183}$ & 73.8 & 74.9 & \textbf{75.4} & 75.2 & 75.1 \\
\bottomrule
\end{tabular}
}
\caption{Ablation study on the efficacy of the upper bound number $K$ using $N$=8.}
\label{supp_table:ablation_upper_K}
\end{table}

% Ablation on save criteria
\begin{table}[t!]
\centering
\resizebox{0.99\linewidth}{!}{%
\begin{tabular}{l|ccccc}
\toprule
Save Criteria & 92 & 183 & 366 & 732 & 1464\\
\midrule
(a) Baseline & 71.9 & 72.5 & 76.0 & 77.4 & 78.7 \\
(b) Every 1 Epoch & 71.8 & 73.5 & 76.3 & 77.5 & 78.6 \\
(c) Every 3 Epochs & 72.2 & 74.7 & 76.8 & 78.0 & 78.6  \\
(d) On Best Epochs (ours) & \textbf{73.4} & \textbf{75.4} & \textbf{77.5} & \textbf{78.6} & \textbf{79.3} \\
\bottomrule
\end{tabular}
}
\caption{Ablation study regarding the efficacy of the save criteria using $N$=8 and $K$=3.}
\label{supp_table:ablation_save_criteria}
\end{table}

\subsection{Criteria for Saving Previous Models}
As described in \cref{subsubsec:save_criteria} of the main paper, one alternative for storing previous models involves saving the model at regular intervals, a method used in \cite{huang2017snapshot, wang2022efficient, yang2022st++}. Thus, we conduct experiments to validate the effectiveness of this approach.
As listed in \cref{supp_table:ablation_save_criteria}, although case (b) exhibits slightly better overall performance than the baseline (a), the difference is marginal. This suggests that storing models at short intervals does not address the coupling problem between the teacher and student networks. In contrast, case (c) shows a significant improvement compared to case (a). Although case (c) functions well, it exhibits limited improvements compared to case (d) which utilizes the proposed save criteria, demonstrating the superiority of the proposed approach. In addition, our approach does not require additional hyperparameter searches to determine appropriate intervals, thereby reducing unnecessary training costs.

\subsection{Loss Weight} \label{supp_sec:loss_weight}
The loss weight $\lambda$ for previous guidance plays a critical role in the overall training process. Since the primary goal of previous guidance is to mitigate confirmation bias--which tends to occur during the early to middle stages of training--its effect is most beneficial before the model becomes overconfident in incorrect predictions. However, in the early phase, the model's predictions are typically unreliable. To address this, we employ a warmup schedule that gradually increases $\lambda$, similar to commonly used learning rate warmup strategies.
Furthermore, as training progresses and the model becomes more stable and accurate, the need for strong regularization from previous guidance naturally decreases. To reflect this, we apply a decay to $\lambda$ in the later stages. Our final schedule (row (e)) allows previous guidance to act most strongly when it is most needed--during the middle of training--while reducing its influence as the model converges.

\cref{supp_table:ablation_loss_weight} presents an ablation study comparing different $\lambda$ scheduling strategies. The fixed (b) and linear decay (c) settings maintain a high weight even in the early stage, where the model's predictions are still unstable. As a result, these configurations show limited improvements, likely due to the guidance based on noisy predictions. In contrast, the linear increase (d) and warmup+decay (e) strategies both incorporate a warmup phase, mitigating this issue and yielding significant performance gains across various label partitions.
While both (d) and (e) outperform other variants, the warmup+decay strategy shows slightly better or comparable results, especially in high-label settings. This indicates that reducing the influence of previous guidance in the late stage, when the model is already well-trained, is beneficial for final performance.
\begin{table}[ht]
\centering
\resizebox{0.99\linewidth}{!}{%
\begin{tabular}{l|ccccc}
\toprule
\textbf{Loss weight ($\lambda$)} & 92 & 183 & 366 & 732 & 1464\\
\midrule
(a) Baseline & 71.9 & 72.5 & 76.0 & 77.4 & 78.7 \\
(b) Fixed (1.0) & 72.1 & 74.2 & 76.7 & 78.2 & 78.6\\
(c) Linear Decay (1.0$\rightarrow$0) & 71.5 & 74.5 & 76.9 & 78.4 & 78.8\\
(d) Linear Increase (0$\rightarrow$1.0) & \textbf{73.6} & \textbf{75.4} & 77.3 & 78.5 & 79.2\\
\midrule
(e) Warmup+Decay (0$\rightarrow$1.0$\rightarrow$0) \; & 73.4 & \textbf{75.4} & \textbf{77.5} & \textbf{78.6} & \textbf{79.3} \\
\bottomrule
\end{tabular}
}
\caption{Ablation study for $\lambda$.}
\label{supp_table:ablation_loss_weight}
\end{table}

\section{Additional Analysis}
\label{supp_sec:additional_analysis}

\begin{figure*}[t]
\centering
\includegraphics[width=0.95\linewidth]{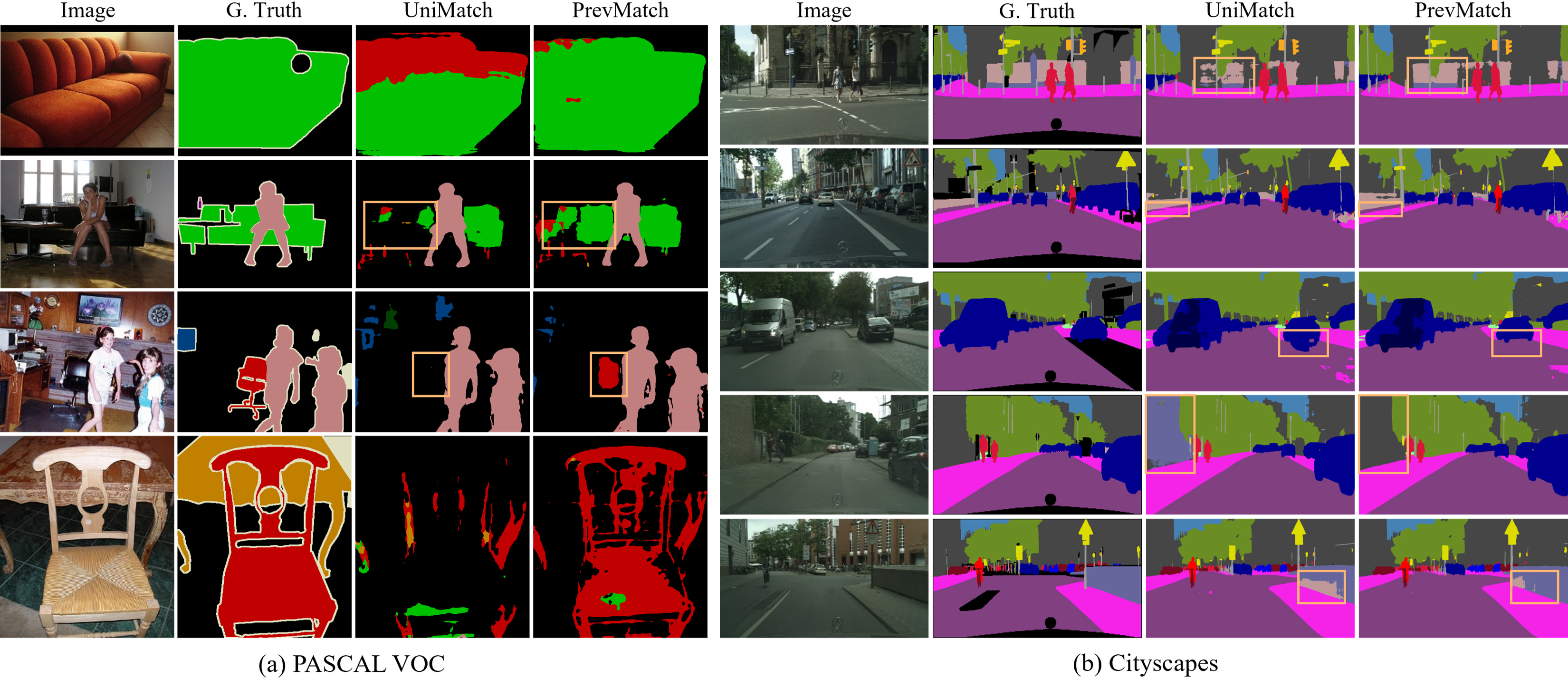}
\caption{Qualitative segmentation results on (a) Pascal VOC and (b) Cityscapes.}
\label{supp_fig:qualitative_results}
\end{figure*}

\begin{table*}[t]
\centering
\resizebox{0.99\linewidth}{!}{%
\begin{tabular}{rccccccccccccccccccccc}
& \rotatebox{75}{\textbf{backgr.}} & \rotatebox{75}{\textbf{airplane}} & \rotatebox{75}{\textbf{bicycle}} & \rotatebox{75}{\textbf{bird}} & \rotatebox{75}{\textbf{boat}} & \rotatebox{75}{\textbf{bottle}} & \rotatebox{75}{\textbf{bus}} & \rotatebox{75}{\textbf{car}} & \rotatebox{75}{\textbf{cat}} & \rotatebox{75}{\textbf{chair}} & \rotatebox{75}{\textbf{cow}} & \rotatebox{75}{\textbf{d.table}} & \rotatebox{75}{\textbf{dog}} & \rotatebox{75}{\textbf{horse}} & \rotatebox{75}{\textbf{m.bike}} & \rotatebox{75}{\textbf{person}} & \rotatebox{75}{\textbf{p.plant}} & \rotatebox{75}{\textbf{sheep}} & \rotatebox{75}{\textbf{sofa}} & \rotatebox{75}{\textbf{train}} & \rotatebox{75}{\textbf{tv}} \\
\midrule
UniMatch & 91 & 84 & 59 & 89 & 71 & 68 & 92 & 80 & 88 & 8 & 88 & 55 & 85 & 85 & 75 & 80 & 54 & 82 & 33 & 80 & 64 \\
 + PrevMatch & 93 & 84 & 61 & 87 & 71 & 68 & 93 & 84 & 89 & 21 & 89 & 57 & 84 & 86 & 77 & 82 & 51 & 82 & 46 & 84 & 58 \\
Gain ($\Delta$) & \cellcolor{green!10}2 & \cellcolor{gray!0}0 & \cellcolor{green!10}2 & \cellcolor{red!10}-2 & \cellcolor{gray!0}0 & \cellcolor{gray!0}0 & \cellcolor{green!5}1 & \cellcolor{green!20}4 & \cellcolor{green!5}1 & \cellcolor{green!50}13 & \cellcolor{green!5}1 & \cellcolor{green!10}2 & \cellcolor{red!5}-1 & \cellcolor{green!10}1 & \cellcolor{green!15}2 & \cellcolor{green!5}2 & \cellcolor{red!15}-3 & \cellcolor{gray!0}0 & \cellcolor{green!50}13 & \cellcolor{green!20}4 & \cellcolor{red!20}-6 \\
\bottomrule
\end{tabular}
}
\caption{Class-wise IoU scores for Pascal VOC using a ResNet-50 encoder.}
\label{supp_table:class_wise_iou_pascal}
\end{table*}

\begin{table*}[t!]
\centering
\resizebox{0.99\linewidth}{!}{%
\begin{tabular}{rccccccccccccccccccc}
& \rotatebox{75}{\textbf{road}} & \rotatebox{75}{\textbf{sidewalk}} & \rotatebox{75}{\textbf{build.}} & \rotatebox{75}{\textbf{wall}} & \rotatebox{75}{\textbf{fence}} & \rotatebox{75}{\textbf{pole}} & \rotatebox{75}{\textbf{t.light}} & \rotatebox{75}{\textbf{t.sign}} & \rotatebox{75}{\textbf{veget.}} & \rotatebox{75}{\textbf{terrain}} & \rotatebox{75}{\textbf{sky}} & \rotatebox{75}{\textbf{person}} & \rotatebox{75}{\textbf{rider}} & \rotatebox{75}{\textbf{car}} & \rotatebox{75}{\textbf{truck}} & \rotatebox{75}{\textbf{bus}} & \rotatebox{75}{\textbf{train}} & \rotatebox{75}{\textbf{m.cycle}} & \rotatebox{75}{\textbf{bicycle}} \\
\midrule
UniMatch & 98 & 82 & 92 & 56 & 60 & 62 & 71 & 79 & 92 & 60 & 95 & 82 & 63 & 95 & 83 & 87 & 79 & 68 & 77 \\
 + PrevMatch & 98 & 84 & 92 & 60 & 63 & 63 & 71 & 80 & 92 & 63 & 95 & 82 & 64 & 95 & 83 & 88 & 81 & 69 & 77 \\
Gain ($\Delta$) & \cellcolor{gray!0}0 & \cellcolor{green!20}2 & \cellcolor{gray!0}0 & \cellcolor{green!30}4 & \cellcolor{green!25}3 & \cellcolor{green!10}1 & \cellcolor{gray!0}0 & \cellcolor{green!10}1 & \cellcolor{gray!0}0 & \cellcolor{green!25}3 & \cellcolor{gray!0}0 & \cellcolor{gray!0}0 & \cellcolor{green!15}1 & \cellcolor{gray!0}0 & \cellcolor{gray!0}0 & \cellcolor{green!10}1 & \cellcolor{green!20}2 & \cellcolor{green!10}1 & \cellcolor{gray!0}0 \\
\bottomrule
\end{tabular}
}
\caption{Class-wise IoU scores for Cityscapes using a ResNet-101 encoder.}
\label{supp_table:class_wise_iou_cityscapes}
\end{table*}

\subsection{Class-wise IoU Scores and Qualitative Evaluation}
\cref{supp_table:class_wise_iou_pascal,supp_table:class_wise_iou_cityscapes} list all category-wise IoU scores. In particular, \cref{supp_table:class_wise_iou_pascal} on Pascal VOC shows that the proposed method achieves notable performance gains for the chair and sofa classes, which were particularly challenging for the UniMatch baseline.
In addition, \cref{supp_table:class_wise_iou_cityscapes} on the Cityscapes dataset shows that the proposed method achieves the largest performance improvements for the wall, fence, and terrain classes, which are the lowest performing among the 19 classes in UniMatch.
In addition to the quantitative results, the qualitative results shown in \cref{supp_fig:qualitative_results} corroborate these findings, revealing consistent improvements in the same categories.
Thus, these results suggest that utilizing previous knowledge helps prevent the catastrophic forgetting problem, even in semi-supervised semantic segmentation scenarios.

\begin{figure*}[t!]
\centering
\includegraphics[width=\linewidth]{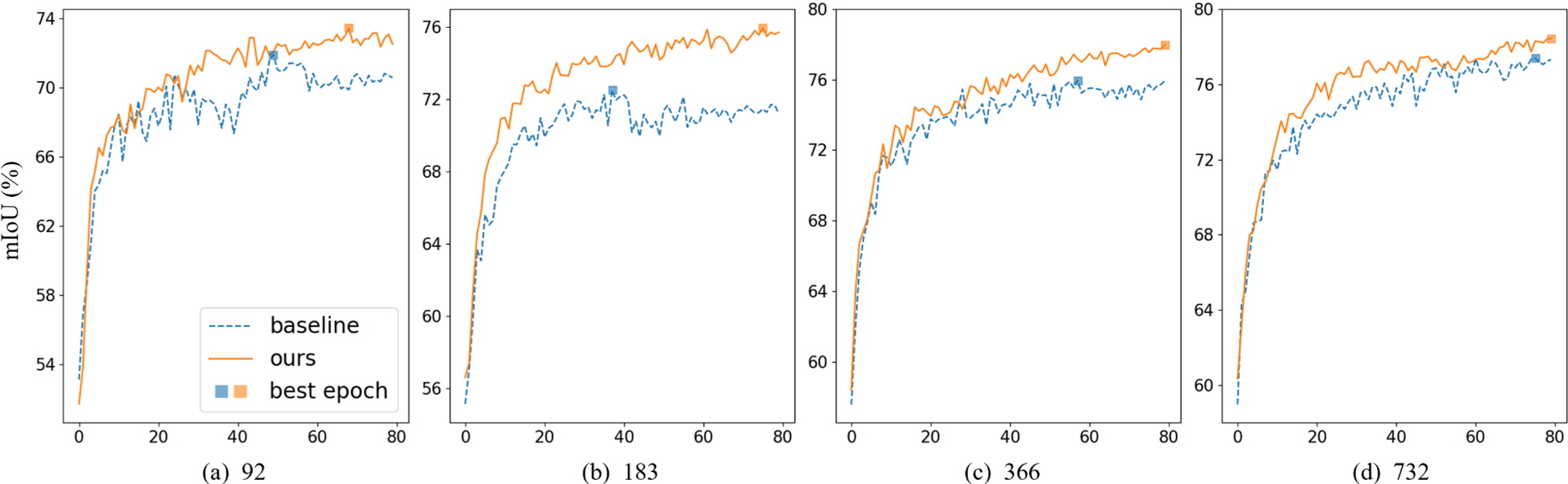}
\caption{Training curves for different label partition settings on Pascal VOC. The X- and Y-axes represent epochs and validation mIoU, respectively. The square symbol ($\blacksquare$) denotes the epoch with the best performance.}
\label{fig:learning_curve}
\end{figure*}

\subsection{Training Stability Across Different Label Partitions}
\label{supp_sec:training_stability}
\cref{fig:learning_curve} illustrates the effect of the proposed method on the changes in the validation scores throughout the training process. In fewer label settings (92 and 183), the UniMatch baseline (blue) exhibits significant fluctuations in terms of performance compared to the proposed method (orange).
Moreover, when considering the epoch that achieves the best performance, the baseline method struggles to converge consistently across epochs and tends to become trapped in local minima prematurely. This issue is particularly pronounced in scenarios with fewer labels. In contrast, the proposed method consistently converges across epochs without significant fluctuations.

Meanwhile, as mentioned in \cref{subsec3:previous_guidance} of the main paper, given the standard and previous guidance, four scenarios can be considered based on whether they are correct or incorrect (standard-previous): (1) correct-correct, (2) correct-incorrect, (3) incorrect-correct, and (4) incorrect-incorrect. Through case (3), the network receives an additional opportunity to be guided in the right direction, away from the wrong one. 
However, model training may also be hindered through case (2), leading to significant fluctuations.
Nevertheless, as shown in \cref{fig:learning_curve}, the consistent outperformance of our method over the baseline across almost all training epochs in different label partitions suggests that the positive effects (i.e., case (3)) of previous guidance outweigh any negative effects (i.e., case (2)).

\section{Limitations and Future Work}
We showed that the proposed method is effective on several benchmark datasets that share the domain between labeled and unlabeled images. However, it has not been investigated in domain adaptation problems that involve domain discrepancies between the labeled and unlabeled data.
This problem poses a more challenging self-training task, a commonly encountered problem in real-world applications.
In future work, we will investigate the capability of temporal knowledge in the domain adaptation problem.
In addition, we observed a phenomenon where significant fluctuations in pseudo-label accuracy for several classes negatively affect generalization ability. For instance, pseudo-label accuracy recovers slightly after a sharp drop in performance; however, the validation score does not.
Although the proposed method has shown that it can mitigate these issues, a more in-depth investigation is required for scenarios involving many classes and imbalanced distributions, such as the COCO and ADE20K datasets.
Therefore, we intend to explore this phenomenon extensively across different classes, in terms of the relationship between training instability and generalization ability.

\end{document}